\documentclass[11pt,a4paper]{article}
\usepackage[hyperref]{acl2017}
\usepackage{times}
\usepackage{subfigure} 

\usepackage{latexsym}
\usepackage{graphicx}
\usepackage{algorithm}
\usepackage{algorithmic}
\usepackage{amsmath}
\usepackage{setspace}
\usepackage{url}

\aclfinalcopy 

\title{A  Deep Network with Visual Text Composition Behavior}

\author{Hongyu Guo \\
  National Research Council Canada\\
  1200 Montreal Road, Ottawa, Ontario, K1A 0R6 \\
  {\tt hongyu.guo@nrc-cnrc.gc.ca} 
	}

\date{}

\begin{document}
\maketitle
\begin{abstract}
While natural languages are compositional, how  state-of-the-art  neural models achieve compositionality  is still unclear. 
We propose a  deep   network, which  not only achieves competitive accuracy for text classification, but also  exhibits  compositional behavior. That is, while creating hierarchical  representations of  a piece of text, such as a sentence, the lower layers of the network distribute their  layer-specific  attention weights to individual words. In contrast, the higher layers compose meaningful phrases and clauses, whose lengths increase as the networks get deeper until fully composing the  sentence. 
\end{abstract}

\section{Introduction}
Deep neural networks leverage task-specific architectures to develop hierarchical representations of the  input,  where  higher level representations are derived from lower level features~\cite{DBLP:journals/corr/ConneauSBL16}.
Such hierarchical representations have visually demonstrated  compositionality in image processing, i.e., pixels  combine to form shapes and then contours ~\cite{DBLP:journals/pami/FarabetCNL13,DBLP:conf/eccv/ZeilerF14}.  
Natural languages are also compositional, i.e., words combine to form  phrases and then sentences. Yet unlike in vision, 
how deep neural models in NLP, which mainly operate
on distributed word embeddings, achieve compositionality, 
 is  still unclear~\cite{DBLP:journals/corr/LiCHJ15,li2016understanding}.

We propose an Attention Gated Transformation (AGT)   network, where  each layer's  feature generation  is  gated by a layer-specific attention mechanism~\cite{bahdanau2014neural}.  Specifically, through distributing its  attention to the original given text, each layer of the networks   
tends to \textit{incrementally} retrieve new words and phrases from the original text. The new knowledge is then  combined with the previous layer's features to create the current layer's representation, thus resulting in composing longer or new phrases and clauses while  creating higher layers' representations  of the text.

Experiments on the Stanford Sentiment Treebank~\cite{Socher2013}  dataset show that the  AGT method not only achieves very competitive accuracy, 
but also  exhibits   compositional behavior via its layer-specific attention. We empirically show that, given a piece of text, e.g., a sentence, the lower layers of the networks  select individual words, e.g, negative and conjunction words \textit{not} and \textit{though}, while the higher layers  aim at composing meaningful phrases and clauses such as negation phrase \textit{not so much}, where the phrase length increases as the networks get deeper  until fully composing the whole sentence. Interestingly,  after  composing the  sentence, 
the compositions of different  sentence phrases 
compete to become the dominating  features of the end task.

\begin{figure}[h]
\caption{An AGT network with three layers.}
	\centering
		\includegraphics[width=2.8in]{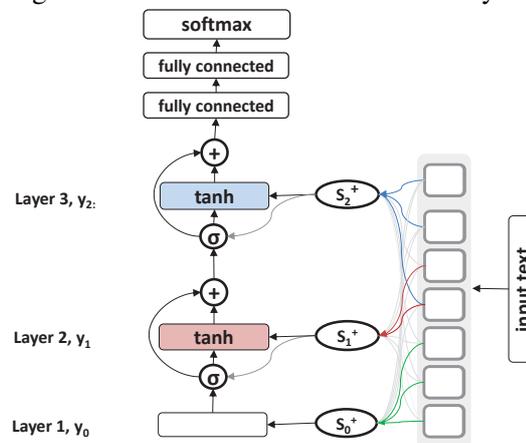}
	\label{fig:schema}
\end{figure}

\section{Attention Gated Transformation Network}
Our AGT network was inspired by the Highway Networks~\cite{DBLP:journals/corr/SrivastavaGS15,Srivastava:2015:TVD:2969442.2969505}, where   each layer is equipped with a \textit{transform gate}.  

\subsection{Transform Gate for Information Flow}
Consider a feedforward neural network with multiple layers. Each layer $l$ typically applies a non-linear transformation $f$ (e.g., $tanh$, parameterized by W$^{f}_{l}$), on its input, which 
is the output of the most recent previous layer (i.e., $y_{l-1}$), to produce its output  $y_{l}$. Here, $l=0$ indicates the first layer and  $y_{0}$ is equal  to the  given input text $x$, namely $y_{0}=x$: 
\begin{align}
y_{l} = f(y_{l-1},W^{f}_{l})
\end{align}
While in a highway network (the left column of Figure~\ref{fig:schema}), an additional non-linear transform gate function $T_{l}$ 
is added to the $l^{th} (l>0)$ layer:
\begin{align}
\label{gate}
y_{l} = f(y_{l-1},W^{f}_{l})· T_{l} + y_{l-1} · (1 - T_{l})
\end{align}
where the function $T_{l}$  expresses how much of the  representation $y_{l}$ is produced by transforming the  $y_{l-1}$ (first term in Equation~\ref{gate}), and how much is just carrying from  $y_{l-1}$ (second term in Equation~\ref{gate}). Here $T_{l}$  is  typically defined as:
\begin{align}
\label{eq3}
T_{l} = \sigma(W_{l}^{t}y_{l-1}+b_{l}^{t})
\end{align}
where $W_{l}^{t}$ is the weight matrix
and $b_{l}^{t}$ the bias vector; $\sigma$ is the non-linear activation function.

With transform gate $T$, the networks learn to  decide if a feature transformation is needed at each layer. 
Suppose $\sigma$ represents a sigmoid function. In such case, the output of $T$ lies between zero and one. Consequently,  
 when the transform gate is one, the networks pass through the transformation $f$ over $y_{l-1}$ and block the pass of input $y_{l-1}$; when the gate is zero, the networks pass through the unmodified  $y_{l-1}$, while the transformation $f$ over $y_{l-1}$ is suppressed.

The left column of Figure~\ref{fig:schema} reflects the  highway networks as proposed by~\cite{Srivastava:2015:TVD:2969442.2969505}. Our AGT method adds  the right two columns of Figure~\ref{fig:schema}. That is, 1) the transform gate $T_{l}$ now is not a function of $y_{l-1}$, but a function of the selection vector $s^{+}_{l}$, which is determined by the attention distributed to the  given input $x$ by the $l^{th}$ layer (will be discussed next), and  2) the  function $f$ takes as input the concatenation of  $y_{l-1}$ and  $s^{+}_{l}$ to create feature representation $y_{l}$. These changes result in an attention gated transformation  when forming  hierarchical representations of the text.

\subsection{Attention Gated Transformation}
\label{agt}
In AGT, the activation of the transform gate at each layer   depends on a layer-specific attention mechanism.  
Formally, given a piece of text $x$, such as a sentence with $N$ words, it can be represented as a matrix $B$ $ \in {\rm I\!R}^{N \times d}$. Each row of the matrix  corresponds to one word, which is represented by a  $d$-dimensional  vector as provided by a learned word embedding  table.  
 Consequently, the   selection vector $s^{+}_{l}$, for the $l^{th}$ layer, is the softmax weighted sum over the $N$  word vectors in $B$:
\begin{align}
\label{eqq}
s^{+}_{l} =\sum^{N}_{n=1} d_{l,n}B[n:n]
\end{align}
with the weight (i.e., attention) $d_{l,n}$   computed as:
\begin{align}
d_{l,n} =\frac{exp (m_{l,n})}{\sum^{N}_{n=1} exp (m_{l,n})}
\end{align}
\begin{align}
m_{l,n} =w^{m}_{l} \text{tanh}( W_{l}^{m}(B[n:n]))
\end{align} 
here, $w_{l}^{m}$ and $W_{l}^{m}$ are the weight vector and weight matrix, respectively. By varying the attention weight $d_{l,n}$,  the  $s^{+}_{l}$ can focus on different rows of the matrix $B$, namely different words of the given text $x$, as illustrated by different color curves connecting to $s^{+}$ in Figure~\ref{fig:schema}. Intuitively, one can consider $s^{+}$  as a learned word selection component: choosing different sets of words of the given text $x$ by distributing its distinct attention. 

Having built one $s^{+}$ for each layer from the given text $x$,  the activation of the transform gate for  layer $l$ $(l > 0)$ (i.e., Equation~\ref{eq3}) is calculated:
\begin{align}
\label{e1}
T_{l} = \sigma(W_{l}^{t}s^{+}_{l}+b_{l}^{t})
\end{align}
To generate  feature representation $y_{l}$, 
 the  function $f$ takes as input  the concatenation of $y_{l-1}$ and $s^{+}_{l}$. That is, Equation~\ref{gate} becomes:
\begin{align}
\label{e2}
    y_{l}= 
\begin{cases}
    s^{+}_{l},&  l = 0\\    f([y_{l-1};s^{+}_{l}],W_{l}^{f})· T_{l} + y_{l-1} · (1 - T_{l})    ,         & l >0 
\end{cases}
\end{align}
where [...;...] denotes concatenation. 
Thus,  at each layer $l$, the gate $T_{l}$ can regulate either passing through    $y_{l-1}$ to form $y_{l}$, or retrieving novel knowledge from the input text $x$ to augment $y_{l-1}$ to create a better representation for $y_{l}$.

 Finally, as depicted in Figure~\ref{fig:schema},  the feature representation of the last layer of the AGT  is fed into two fully connected layers followed by a softmax function to produce a distribution over the  possible target classes. 
For training, we use multi-class cross entropy loss. 

Note that,  Equation~\ref{e2} indicates that the representation $y_{l}$ depends on both  $s^{+}_{l}$ and $y_{l-1}$. 
In other words, although Equation~\ref{e1} states that the gate activation at layer $l$ is computed by $s^{+}_{l}$,  the gate activation is also affected by $y_{l-1}$, which embeds the information from the layers below $l$.

Intuitively, the AGT networks are encouraged to  consider new words/phrases from the input text at higher layers. Consider the fact that the $s_{0}^{+}$  at the bottom layer  of the AGT only deploys a  linear transformation of  the bag-of-words features. If no new words are used at higher layers of the networks, it will be challenge for the AGT to sufficiently explore   different combinations of word sets of the given text, which may be important for building an accurate classifier.  In contrast, through  tailoring its attention for new words  at different layers, the AGT enables the words selected by a layer to be effectively combined with  words/phrases selected by its previous layers to  benefit the accuracy of the classification task (more discussions are presented in Section~\ref{discussion}).

\section{Experimental Studies}
\label{expr}
\subsection{Main Results}
\label{setup}
The    Stanford Sentiment Treebank  data  contains  11,855  movie reviews~\cite{Socher2013}. We use the same splits for training, dev, and test data as in~\cite{Kim14} to predict the fine-grained 5-class sentiment categories of the sentences.
For  comparison purposes, following~\cite{Kim14,DBLP:journals/corr/KalchbrennerGB14,DBLP:journals/corr/LeiBJ15}, we trained  the models  using both phrases and sentences, but only evaluate sentences at test time. 
Also,  we initialized all of the word embeddings~\cite{cherry2015unreasonable,chen2015representation} using the  300 dimensional pre-trained  vectors from GloVe~\cite{pennington2014glove}. We learned 15 layers with 200 dimensions each,  which requires us to project the 300 dimensional word vectors; we implemented this using a linear transformation, whose weight matrix and bias term are shared across all words, followed by a $\tanh$ activation. For optimization, we used Adadelta~\cite{DBLP:journals/corr/abs-1212-5701}, with  learning
rate of 0.0005, mini-batch of  50,  transform gate bias of 1, and dropout~\cite{Srivastava:2014} rate of 0.2. 
All  these  hyperparameters were determined through experiments on the  validation-set.

\renewcommand{\arraystretch}{0.6}
\begin{table}[h]
  \centering
\begin{tabular}{l|c}\hline
AGT & 50.5\\\hline 
high-order CNN& \textbf{51.2}\\
tree-LSTM& 51.0\\ 
DRNN& 49.8 \\
PVEC& 48.7 \\
DCNN& 48.5 \\
DAN &48.2 \\
CNN-MC& 47.4 \\
CNN & 47.2\\
RNTN& 45.7\\
NBoW&  44.5 \\
RNN &43.2 \\
SVM & 38.3 \\
 \hline
\end{tabular}
  \caption{Test-set accuracies obtained; results except the AGT are drawn  from~\cite{DBLP:journals/corr/LeiBJ15}.}
  \label{tab:accuracy:sen}
\end{table}

\begin{figure}[H]
\caption{Soft attention distribution (top) and phrase length distribution (bottom) on the test set. }
	\centering
		\includegraphics[width=2.01238in]{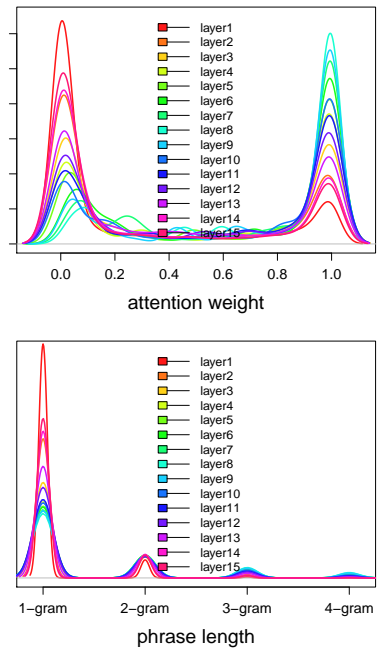}
	\label{fig:merge}
\end{figure}

Table~\ref{tab:accuracy:sen} presents the test-set accuracies obtained by different  strategies. Results in Table~\ref{tab:accuracy:sen} indicate that the AGT method  achieved very competitive accuracy (with 50.5\%), when compared to the state-of-the-art results    obtained by the  tree-LSTM (51.0\%)~\cite{DBLP:journals/corr/TaiSM15,zhu2015long} and high-order CNN approaches (51.2\%)~\cite{DBLP:journals/corr/LeiBJ15}. 

 Top subfigure in Figure~\ref{fig:merge} depicts the distributions of the attention weights created by different layers on all  test data, where the attention weights of all words in a sentence, i.e.,  $d_{l,n}$ in Equation~\ref{eqq}, are  normalized to the range between 0 and 1 within the sentence.  
The figure indicates that  AGT generated very spiky attention distribution. That is, most of the attention weights are either 1 or 0. Based on these  narrow, peaked bell curves formed by the normal distributions for the attention weights of 1 and 0,  we here consider a word has been \textit{selected} by the networks if its attention weight is larger than 0.95, i.e., receiving more than 95\% of the full attention, and a phrase has been \textit{composed and selected} if a set of consecutive words all have been selected. 

In the bottom subfigure of Figure~\ref{fig:merge} we present the distribution of the phrase lengths  on the test set. 
 This figure indicates that the middle layers of the networks  e.g.,  8$^{th}$ and 9$^{th}$, have longer  phrases  (green and blue curves) than others, while the layers at the two ends contain shorter phrases (red and pink curves).

\begin{figure}
\caption{Transform gate activities of the test-set (top) and the first  sentence in Figure~\ref{fig:sen1} (bottom).}
	\centering
		\includegraphics[width=2.124in,height=2.7in]{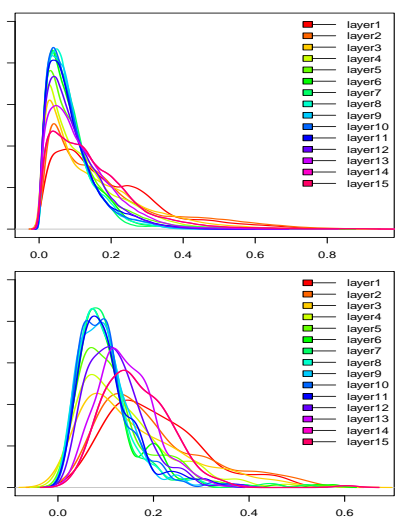}
	\label{fig:gates}
\end{figure}

\begin{figure*}[h]
\caption{Three  sentences from the test set and their attention received from the 15 layers (L1 to L15). }
	\centering
		\includegraphics[width=6.59535905024in,height=3.454in]{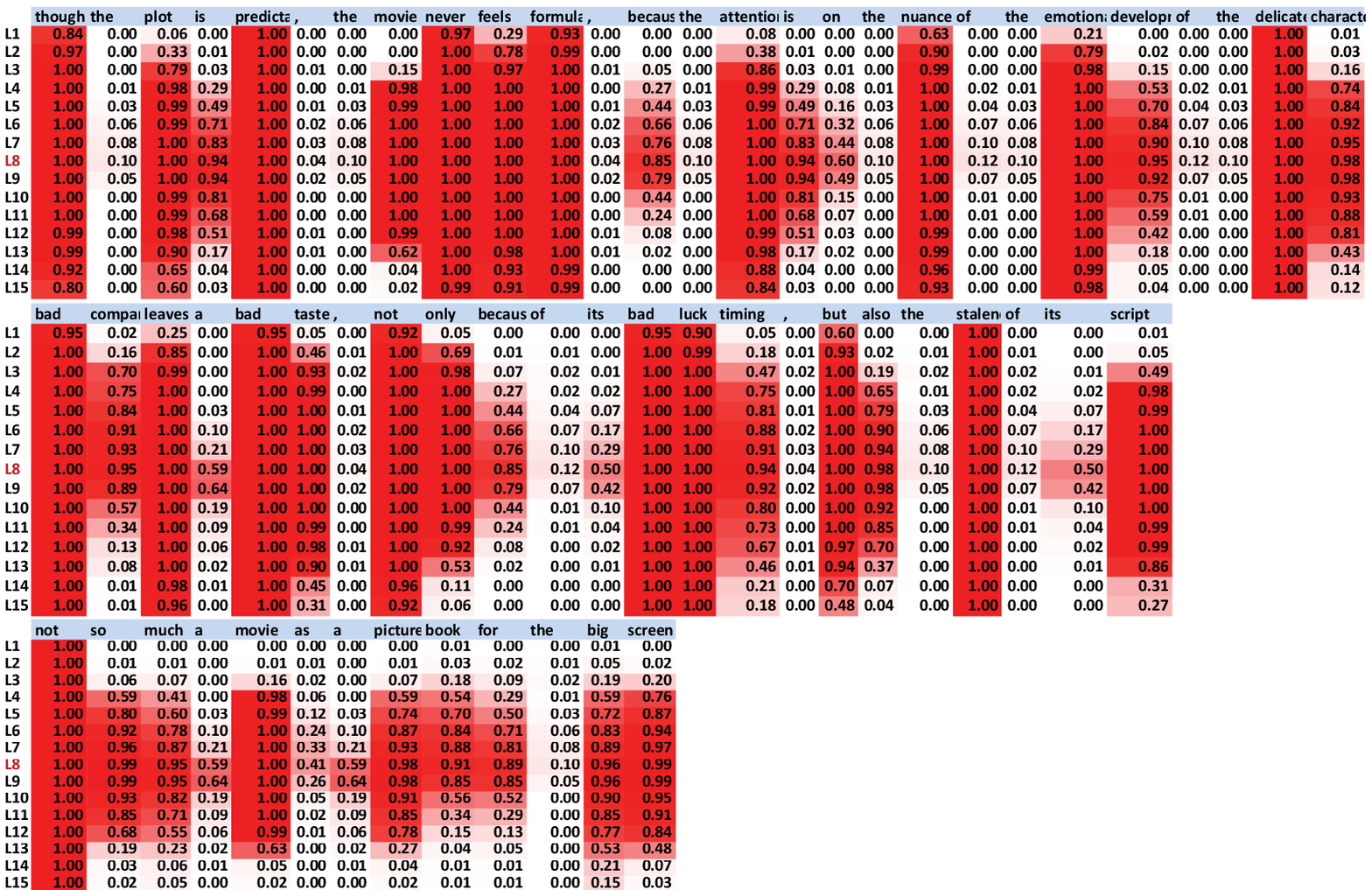}
	\label{fig:sen1}
\end{figure*}

In Figure~\ref{fig:gates}, we also presented the transform gate activities on all test sentences (top) and that of the first example sentence in Figure~\ref{fig:sen1} (bottom). These curves suggest that the transform gates  at the middle layers (green and blue curves) tended to be close to zero, indicating the pass-through of lower layers' representations.  On the contrary, the gates at the two ends (red and pink curves) tended to be away from zero with large tails, implying the retrieval of new knowledge from the input text. These  are consistent with the results   below.

Figure~\ref{fig:sen1}  presents three sentences with various lengths from the test set, with  the attention weights  numbered and  then highlighted in heat map. 
 Figure~\ref{fig:sen1} suggests  that  the lower layers of the networks selected individual words, while the higher layers aimed at phrases.  
For example, the first and second layers seem to \textit{select} individual words carrying strong sentiment (e.g.,  \textit{\textcolor{blue}{predictable}}, \textit{\textcolor{blue}{bad}}, \textit{\textcolor{blue}{never}} and \textit{\textcolor{blue}{delicate}}), and conjunction and negation words (e.g., \textit{\textcolor{blue}{though}} and \textit{\textcolor{blue}{not}}). 
Also, meaningful phrases were \textit{composed and selected} by later layers, such as \textit{\textcolor{blue}{not so much}}, \textit{\textcolor{blue}{not only... but also}}, \textit{\textcolor{blue}{bad taste}}, \textit{\textcolor{blue}{bad luck}}, \textit{\textcolor{blue}{emotional development}}, and \textit{\textcolor{blue}{big screen}}.  
In addition, in the middle layer, i.e., the 8$^{th}$ layer, the whole sentences were composed by filtering out uninformative words, resulting in concise versions, as follows (selected words and  phrases are highlighted in color blocks).  
\begin{quote}
1) \textit{\textcolor{blue}{though }} \textcolor{red}{plot} \textit{\textcolor{blue}{predictable}} \textcolor{red}{movie never feels formulaic} \textit{\textcolor{blue}{attention}}  \textcolor{red}{nuances} \textit{\textcolor{blue}{emotional development}} \textcolor{red}{delicate characters} \\
2) \textit{\textcolor{red}{bad company leaves}} \textcolor{blue}{bad taste} \textit{\textcolor{red}{not only}} \textcolor{blue}{bad luck} \textit{\textcolor{red}{but also}} \textcolor{blue}{staleness} \textit{\textcolor{red}{script}} \\
3) \textit{\textcolor{red}{not so much}} \textcolor{blue}{movie}  \textit{\textcolor{red}{picture}} \textcolor{blue}{big screen} 
\end{quote}

Interestingly, if relaxing the word  selection criteria, e.g., including words receiving more than the median, rather than 95\%, of the full attention, the  sentences  recruited more conjunction and modification words, e.g., \textit{\textcolor{blue}{because}}, \textit{\textcolor{blue}{for}}, \textit{\textcolor{blue}{a}}, \textit{\textcolor{blue}{its}} and \textit{\textcolor{blue}{on}}, thus becoming more readable and fluent: 
\begin{quote}
1) \textit{\textcolor{blue}{though}} \textcolor{red}{plot is predictable} \textit{\textcolor{blue}{movie never feels formulaic}} \textcolor{red}{because} \textit{\textcolor{blue}{attention is on}} \textcolor{red}{nuances} \textit{\textcolor{blue}{emotional development}} \textcolor{red}{delicate characters} \\
2) \textit{\textcolor{blue}{bad company leaves a bad taste}} \textcolor{red}{not only because} \textit{\textcolor{blue}{its bad luck timing}} \textcolor{red}{but also} \textit{\textcolor{blue}{staleness}} \textcolor{red}{its script} \\
3) \textit{\textcolor{blue}{not so much a movie}} \textcolor{red}{a picture book for} \textit{\textcolor{blue}{big screen} } 
\end{quote}
Now, consider the  AGT's compositional behavior for  a specific sentence, e.g., the last sentence in Figure~\ref{fig:sen1}. The first layer solely \textit{selected} the word \textit{\textcolor{blue}{not}} (with  attention weight of 1 and all other words with  weights close to 0), but the $2^{nd}$ to $4^{th}$ layers  gradually pulled out new words \textit{\textcolor{blue}{book}}, \textit{\textcolor{blue}{screen}} and \textit{\textcolor{blue}{movie}} from the  given  text. Incrementally, the $5^{th}$ and $6^{th}$ layers further \textit{selected}  words to form phrases \textit{\textcolor{blue}{not so much}}, \textit{\textcolor{blue}{picture book}}, and \textit{\textcolor{blue}{big screen}}. Finally, the $7^{th}$ and $8^{th}$ layers  added some conjunction  and quantification words \textit{\textcolor{blue}{a}} and \textit{\textcolor{blue}{for}} to make the sentence more fluent. This recursive composing process resulted in the sentence ``\textit{\textcolor{blue}{not so much a movie a picture book for big screen}}''.

Interestingly, Figures~\ref{fig:sen1} and ~\ref{fig:merge} also imply that,  after  composing the sentences by the middle layer,  the AGT networks shifted to  re-focus on  shorter phrases and informative words. 
 Our analysis on the transform gate activities suggests that, during this re-focusing stage  the compositions of  sentence phrases competed to each others, as well as to the whole sentence composition, for the dominating task-specific features  to represent the text.

\subsection{Further Observations}
 \label{discussion}
As discussed at the end of Section~\ref{agt}, intuitively, including new words at different layers allows the networks to  more effectively explore different combinations of  word sets of the given text than that of using all words only at the bottom layer of the networks.
Empirically, we observed that, if with only $s^{+}_{0}$ in the AGT network, namely removing $s^{+}_{i}$ for $i > 0$, the test-set  accuracy dropped from 50.5\% to 48.5\%. In other words, transforming a linear combination of the bag-of-words features was insufficient for obtaining sufficient accuracy for the classification task. For instance, if being augmented with two more selection vectors $s^{+}_{i}$, namely removing $s^{+}_{i}$ for $i > 2$,  the AGT was able to improve its accuracy to 49.0\%.

Also, we observed that the AGT networks tended to select informative words at the lower layers. This may be caused by the  recursive form of Equation~\ref{e2}, which suggests that the words retrieved by $s^{+}_{0}$ have more chance to combine with and influence the selection of other feature words. In our study, we found that, for example, the top 3 most frequent words selected by the first layer of the AGT networks were all negation words: \textit{n't, never}, and \textit{not}. These are important words for sentiment classification~\cite{zhu2014empirical}.

In addition, like the transform gate in the Highway networks~\cite{DBLP:journals/corr/SrivastavaGS15} and the forget gate in the LSTM~\cite{Gers:2000:LFC:1121912.1121915}, the attention-based transform gate in the AGT  networks is sensitive to its bias initialization. We found that initializing the bias to one  encouraged the compositional behavior of the AGT networks.

\section{Conclusion and Future Work}
 We have presented a novel deep network. 
It  not only achieves very competitive accuracy for text classification, but also  exhibits interesting text compositional behavior,  
 which may shed light on  understanding how neural models  work in NLP tasks. 
In the future, we aim to apply the AGT networks to incrementally generating natural text~\cite{guo2015deep,hu2017controllable}.

\bibliography{reference}
\bibliographystyle{acl_natbib}

\end{document}